\documentclass[aapm,mph,amsmath,amssymb,reprint]{revtex4-2}

\usepackage{graphicx}
\usepackage[table]{xcolor}
\newlength{\cellw}
\newcommand{\has}{\rule[-0.9ex]{\cellw}{3.1ex}}
\newcommand{\qual}{{\setlength{\fboxsep}{0pt}\setlength{\fboxrule}{0.5pt}\framebox[\cellw]{\rule[-0.6ex]{0pt}{2.4ex}}}}
\newcommand{\no}{\hspace{\cellw}}
\newcommand{\hd}[1]{\makebox[\cellw]{#1}}
\usepackage{booktabs}
\usepackage{siunitx}
\usepackage{tikz}
\usetikzlibrary{arrows.meta, positioning, fit, backgrounds, calc}
\usepackage[colorlinks=true,allcolors=blue]{hyperref}

\newcommand{\datasetdoi}{10.5281/zenodo.22139642}
\newcommand{\datasetversiondoi}{10.5281/zenodo.22797014}
\newcommand{\datasetversion}{v11}
\newif\ifreview \reviewfalse
\newif\ifcaptionlist \captionlistfalse
\newif\ifsupplement \supplementtrue
\newcommand{\doiurl}[1]{\url{https://doi.org/#1}}
\newcommand{\consortium}{\ifreview a student research consortium\else OpenSpineConsortium\fi}

\newcommand{\Toolkit}{\ifreview An open-source measurement toolkit\else OpenSpineToolkit, a unit-tested open-source package\fi}

\newsavebox{\fitbox}
\newcommand{\maxwidth}[1]{%
  \sbox{\fitbox}{#1}%
  \ifdim\wd\fitbox>\linewidth
    \resizebox{\linewidth}{!}{\usebox{\fitbox}}%
  \else
    \usebox{\fitbox}%
  \fi
}

\newsavebox{\capbox}
\newcommand{\fitfloat}[2]{
  \sbox{\capbox}{\parbox{\linewidth}{\caption{#2}}}%
  \sbox{\fitbox}{#1}%
  \ifdim \wd \fitbox>\linewidth
    \sbox{\fitbox}{\resizebox{\linewidth}{!}{\usebox{\fitbox}}}%
  \fi
  \dimen0=\dimexpr \ht \fitbox+\dp \fitbox \relax
  \dimen2=\dimexpr \textheight-\ht \capbox-\dp \capbox-\baselineskip \relax
  \ifdim \dimen0>\dimen2
    \sbox{\fitbox}{\resizebox{!}{\dimen2}{\usebox{\fitbox}}}%
  \fi
  \usebox{\fitbox}\par \usebox{\capbox}%
}

\begin{document}
\setlength{\dbltextfloatsep}{8pt plus 2pt minus 2pt}
\setlength{\textfloatsep}{8pt plus 2pt minus 2pt}
\setlength{\dblfloatsep}{8pt plus 2pt minus 2pt}
\setlength{\floatsep}{8pt plus 2pt minus 2pt}
\setlength{\intextsep}{8pt plus 2pt minus 2pt}
\setlength{\abovecaptionskip}{1.3pt plus 1pt minus 1pt}
\setlength{\belowcaptionskip}{0pt}

\title{CTSpinoPelvic1K: spine, pelvis, ribs and femora in one coordinate frame,
       annotated for lumbosacral transitional anatomy}

\ifreview
\else
\author{Gregory Schwing}\email{gregory.schwing@med.wayne.edu}
\affiliation{Department of Surgery, Detroit Medical Center and Wayne State University, Detroit, Michigan, USA}
\author{Ashley Schehr}
\author{Annika Tekumulla}
\author{Margret Khoushi}
\author{Ryan Christian}
\author{Dane Hubers}
\author{Faris Mahjoub}
\author{Hassan Saad}
\author{Mia Sooch}
\author{Sathyagopal Siddapureddy}
\author{Michael McLellan}
\author{Jerick Kim}
\affiliation{School of Medicine, Wayne State University, Detroit, Michigan, USA}
\author{Miraziz Ismoilov}
\author{Nizar Alnabahneh}
\affiliation{Department of Radiology, Detroit Medical Center and Wayne State University, Detroit, Michigan, USA}
\fi

\date{\today}

\begin{abstract}
\textbf{Purpose:} The gold standard for naming a vertebra at the lumbosacral junction is counting caudally from C2 on whole-spine imaging. In practice, a lumbar surgical case is planned on lumbar-only imaging (T12 to S1) without C2, making the gold standard method impossible.
Abdominopelvic computed tomography (CT) holds the lumbar-only imaging span plus the lowest ribs and pelvis,
making it a reasonable substitute for the preoperative view. When a
lumbosacral transitional vertebra (LSTV) alters the count, however, the local anatomy is much more ambiguous. A person with four lumbar vertebrae may have an L1 bearing a lumbar rib or an
L5 assimilated to the sacrum, and a person with six may have an L6, a T12 with aplastic
ribs, or a lumbarized S1. CTSpinoPelvic1K permits asking whether local morphology
can resolve that ambiguity without counting caudally from C2. Two public label sets already covered this imaging, CTSpine1K's vertebrae and CTPelvic1K's pelvis, on the same colonography
patients. This release joins them in one dataset of 802 CT records and adds ribs, femora and surgical hardware. It also introduces classes for L6, T13
and lumbar ribs, a class vocabulary that describes more of the anomalies at this junction than any existing public CT collection.

\textbf{Acquisition and Validation Methods:} Records pair CTSpine1K and CTPelvic1K masks on each patient's bone-richest TCIA acquisition, and the release carries the series keys that map every mask to its volume. Validation covered geometric invariants (802/802 pass), rib--vertebra incidence across 5{,}753 ribs (0.035\% offset), and spinopelvic measures matching published values.

\textbf{Data Format and Usage Notes:} NIfTI image/label pairs with patient-grouped LSTV-stratified five-fold splits and a loader; archived at \doiurl{\datasetdoi}.

\textbf{Potential Applications:} Classifying a vertebra from local features;
updating cadaveric morphometry;
spinopelvic assessment; opportunistic screening; imaging-based prevalence of anatomical variants in an asymptomatic cohort; and spinopelvic biomechanical modelling from patient-matched prone and supine acquisitions.
\emph{Limitations:} thoracic ground truth is field-of-view limited; postural
angles are non-standing; no held-out test set; ribs are triaged-review pseudolabels;
Castellvi grades are two-reader consensus on 33 records.
\end{abstract}

\maketitle

\section{Introduction}

The spine typically has seven cervical, twelve thoracic, and five lumbar vertebrae. A
vertebra is classified by where it falls in that sequence. Transitional variants sit at the
two ends of the lumbar column, with the thoracolumbar transitional vertebra (TLTV) above and the
lumbosacral transitional vertebra (LSTV) below. Both transitional variants disturb identification by changing what the border vertebra
and its ribs \emph{look like}. For example, a thoracolumbar border vertebra may carry a full rib, a hypoplastic stump, or none at all, and
a lumbosacral one may be partly or completely assimilated to the sacrum
(\emph{sacralization}), with an anteriorly wedged body, a reduced disc beneath it and
hypoplastic facet joints, or separated from it (\emph{lumbarization}), with a
squared body, lumbar-type facets and a full-height disc~\cite{koninwalz2010}. In addition, they
change how many vertebrae each region holds: eleven or thirteen thoracic and four or six lumbar.

Together, these transitional variants leave the vertebra at each border without a definitive name. The one formal nomenclature at the lumbosacral junction, Castellvi's, grades the morphology of the transverse process and says nothing about whether the segment is L5 or S1~\cite{castellvi1984,koninwalz2010}; every imaging landmark proposed for numbering has proved unreliable in transitional cases~\cite{carrino2011,farshad2015,tureli2014}; and even a count from C2 yields a position, which a convention then converts to a name. Each of the readings above yields the same count (Fig.~\ref{fig:anchors}), and the morphology that would
separate them is not always decisive to human readers. Whether it differs
enough for a trained classifier to separate them, or at least to assign each reading a probable classification,
is the question this release aims to answer. The gold standard for numeration in
transitional anatomy is whole-spine imaging, counting caudally from C2~\cite{lian2018},
which determines how many vertebrae and how many rib pairs the column holds. No
spine-limited acquisition can supply either count. LSTV is common, reported at 10--29\% in the general
population~\cite{hanhivaara2022} and in 16.3\% of a whole-spine CT
series~\cite{nagata2025}, and wrong-level surgery is its most serious consequence.

Existing public collections cannot address this, and not for lack of size
(Table~\ref{tab:priorart}). Prior spine datasets omit the pelvis and pelvic datasets do not number
the vertebrae. TotalSegmentator~\cite{totalsegmentator} has no class for a sixth lumbar vertebra or for a rib on a
lumbar vertebra, and VerSe~\cite{verse2021}, the one collection with an L6 class, stops at the sacrum and
carries no ribs. As such, none of these collections can record a spine with six lumbar vertebrae together with both of its
borders, and a segmenter trained on them must shift the
whole column by a level or absorb a vertebra into its neighbor when it encounters an enumeration anomaly. This problem has now been measured
on CT, where prior labeling methods assigned every vertebra correctly in 77\% of subjects and
an anomaly-aware extension of SPINEPS~\cite{spineps2025} raised that to
99\%~\cite{veridah2026}. Its weights label L6 and T13 on CT, but its 1{,}536-scan CT cohort is in-house and
unreleased, so what exists is a capability, not a dataset. 

When considering the intraoperative side, LevelCheck registers the intraoperative
radiograph to the preoperative CT and projects the CT's vertebral labels onto
it~\cite{otake2012,lo2015,desilva2016}. Those labels, however, are placed by hand and verified by
the surgeon~\cite{otake2012}, so the method assumes a correctly labeled CT and does not supply
one itself. As such, a dataset carrying the anomaly classes is vital for automating that step.

CTSpinoPelvic1K places the spine, pelvis, ribs and femora in one coordinate frame and classifies each structure an enumeration anomaly produces, including a sixth lumbar vertebra and lumbar ribs. A scheme without those classes is unable to adequately describe anatomic anomalies.

The project began with the question whether a vertebra can be named from the anatomy on the images a lumbar surgical case is actually planned on. By guideline those images are lumbar-only~\cite{acr2021,nass2013,acrct2022,acrmri2023} and whole-spine coverage is reserved for trauma with an identified
injury~\cite{sixta2012,tqip2018} and for deformity
radiographs. An abdominopelvic CT runs from the diaphragm to the pelvic floor, so it captures
that span with the lowest ribs and the whole pelvis, while lacking C2. Every record
comes from one prospective trial protocol, ACRIN 6664~\cite{colonog,tcia}, which scanned 802 patients aged 50 and over, supine and prone, on multidetector CT at 15 centers, on five scanner
vendors and nine models, with manufacturer, model and reconstruction kernel recorded per
record. To our knowledge, it is the largest spine-and-pelvis annotated CT cohort acquired under a
single protocol. The comparators, by contrast, pool collections, are multi-site by design, or are
routine clinical scans under many protocols (Table~\ref{tab:priorart}). With the protocol for this study fixed, scanner effects on a model become measurable rather than confounded. The downside is the restriction to an abdominal field of view (FOV), in which only the lowest thoracic levels are present
(Sec.~\ref{sec:applications}).

The gap was evident. CTPelvic1K~\cite{ctpelvic1k} and CTSpine1K~\cite{ctspine1k} each annotated the COLONOG
collection under radiologist supervision for the same patients, with one annotating the pelvis and the other annotating the vertebrae. The two had remained separate, although joining them needs no new imaging and no new radiologist. It was not a file merge, because each annotation had to be traced back to the series it was
drawn on (Sec.~\ref{sec:sources}). Once completed, that merge yielded a combined spine-and-pelvis frame
for 802 patients, and the bones neither set had, ribs, femora and hardware, could be
annotated against it rather than from scratch.

\subsection{Vertebra labeling when the scan cannot settle the count}

The main limitation is that no scan in this dataset contains C2, so the gold standard
conventional count cannot be performed. This is a property of abdominal imaging rather than of the
annotation itself. A thirteenth thoracic vertebra and an L1 with a lumbar rib are different phenotypes,
distinguished as separate subtypes in cadaveric classifications of the thoracolumbar
junction~\cite{duplessis2018,poolman2023}, which separate them on quantitative shape rather
than on position. The first is a thirteenth rib-bearing vertebra above five lumbar vertebrae,
so the column has 25 presacral vertebrae; the second is a rudimentary rib on a lumbar-type L1
in a column with the standard 24. Counting distinguishes neither, because four lumbar vertebrae
can mean a lumbar rib, a cranially shifted T13 or an L5 assimilated to the sacrum, and six can
mean an aplastic twelfth rib or a sixth lumbar vertebra. A \emph{stump rib} is a hypoplastic rib that is a key indicator of a
thoracolumbar transitional vertebra. On whole-spine CT scans, rib anomalies travel with the lumbosacral border and hypoplastic
twelfth ribs occur with sacralization and lumbar ribs with lumbarization.~\cite{nagata2025}

In this release, every vertebra carries the identifier its radiologist-sourced annotation
assigned as the ground truth, corrected where the source was wrong. What is measured, however,
is the bone and not its place in a sequence: body height, end-plate width, canal width and
depth, pedicle width, wedge ratio, transverse-process span and its distance from the ala, and
the length of the lowest rib as a fraction of the one above it. Each describes a vertebra in
its own right. None of them shifts with the name a reader gives the junction, so cases that
disagree about the name remain comparable, and a criterion built on them is not a count under
another name.

A vertebra can be named from its own morphology, and that a level has a characteristic
shape is long established: pedicle width and height change systematically from the thoracic to
the lumbar spine,\cite{zindrick1987} vertebral body, end-plate and canal dimensions differ by
level,\cite{panjabi1991,panjabi1992,berry1987} and the transverse process and the iliolumbar
ligament mark the last lumbar vertebra independently of any
count.\cite{hughes2006,koninwalz2010} Cadaveric series of the thoracolumbar junction classify
a border vertebra by quantitative shape alone.\cite{duplessis2018,poolman2023} If those
differences hold at the boundary, the phenotype is decidable locally, and on this dataset it is. Separating T12 from L1 on shape
alone gives an area under the curve of 0.990 and 97.3\% accuracy over 1485 vertebrae. This was calculated using a
logistic regression on five released per-level measures and six ratios between them. Each
measure was divided by the patient's median across their levels, so that size, which
separates thoracic from lumbar trivially and says nothing at the junction, is removed. Folds
are grouped by case and the curve is taken on pooled out-of-fold scores
(\texttt{test\_morphometric\_separability.py} in the released code). Schinz et al.\ reach
the same conclusion from the other side: on 1{,}242 whole-thoracolumbar CTs a shape-based
labeling of the junction matched nerve morphology in every case, against 92.6--97.2\% for
counting- and rib-based rules.\cite{schinz2026}


\begin{table*}[t]
\caption{Public CT collections at the lumbosacral junction. Black, class present; outlined,
graded only to exclude it~\cite{versedata2021}.}
\label{tab:priorart}
\begin{ruledtabular}
\setlength{\cellw}{\dimexpr(\linewidth-50mm)/9\relax}%
\begin{tabular}{@{}l@{}r@{\hspace{4mm}}l@{}}
Collection & Scans & \hd{Count}\hd{L6}\hd{Sacrum}\hd{S1}\hd{Pelvis}\hd{Ribs}\hd{L.\,rib}\hd{Femora}\hd{Grade} \\ \colrule
CTSpine1K~\cite{ctspine1k} & 1{,}005 & \has\has\no\no\no\no\no\no\no \\
CTPelvic1K~\cite{ctpelvic1k} & 1{,}184 & \no\no\has\no\has\no\no\no\no \\
VerSe~\cite{verse2021} & 374 & \has\has\no\no\no\no\no\no\qual \\
RibSeg~v2~\cite{ribsegv2} & 660 & \no\no\no\no\no\has\no\no\no \\
TotalSegmentator~\cite{totalsegmentator} & 1{,}204 & \has\no\has\has\has\has\no\has\no \\
\textbf{CTSpinoPelvic1K} & 802 & \has\has\has\no\has\has\has\has\has \\
\end{tabular}
\end{ruledtabular}
\end{table*}

\section{Acquisition and Validation Methods}

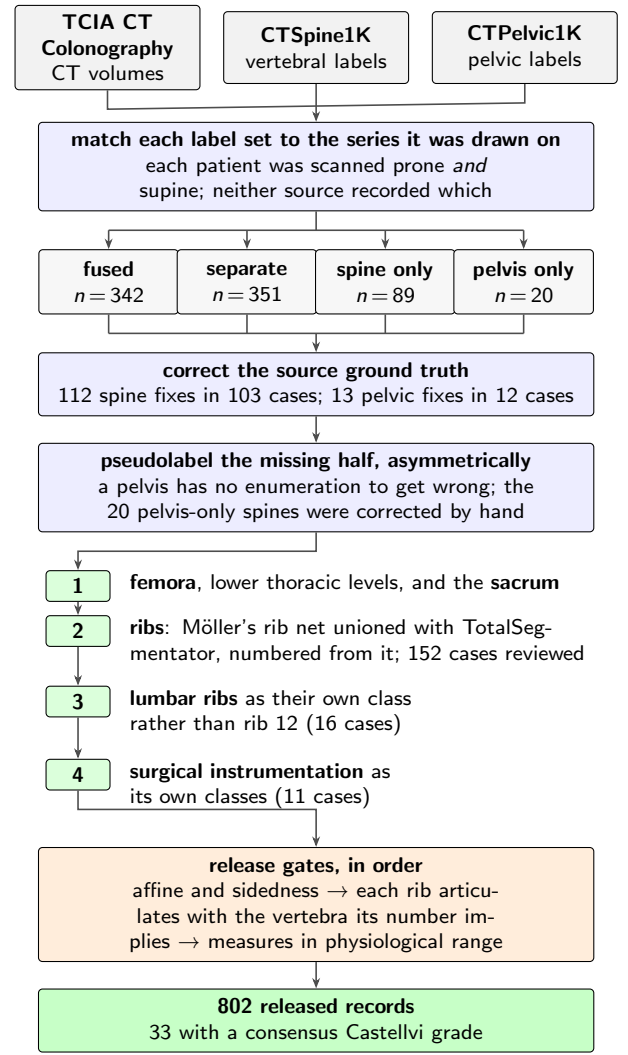
\begin{figure}[t]
\centering
\linespread{1}\selectfont
\begin{tikzpicture}[
  x=1mm, y=1mm, font=\sffamily\fontsize{8}{9.6}\selectfont, node distance=3.5mm,
  bx/.style={draw, rounded corners=1.5pt, align=center, text width=#1, inner sep=3.5pt},
  src/.style={bx=22mm, fill=black!5, minimum height=9.5mm},
  proc/.style={bx=71mm, fill=blue!7},
  rec/.style={bx=16mm, fill=black!3, minimum height=8.5mm},
  ver/.style={bx=6mm, fill=green!14, inner ysep=3pt},
  det/.style={align=left, text width=62mm, inner sep=0pt},
  gate/.style={bx=71mm, fill=orange!14},
  fin/.style={bx=71mm, fill=green!22},
  ar/.style={-{Stealth[length=4pt]}, semithick, black!70},
  ln/.style={semithick, black!70},
]
\node[src] (spine) at (0, 0) {\textbf{CTSpine1K}\\vertebral labels};
\node[src, left=3mm of spine]  (tcia)   {\textbf{TCIA CT Colonography}\\CT volumes};
\node[src, right=3mm of spine] (pelvic) {\textbf{CTPelvic1K}\\pelvic labels};
\node[proc, below=5mm of spine] (cross) {\textbf{match each label set to the series it was drawn on}\\
  each patient was scanned prone \emph{and} supine; neither source recorded which};
\draw[ln] (tcia.south) -- ($(tcia.south)+(0,-2.5mm)$) -- ($(pelvic.south)+(0,-2.5mm)$) -- (pelvic.south);
\draw[ar] (spine.south) -- (cross.north);
\node[rec, anchor=north] (fused) at ($(cross.south)+(-27.45mm,-5mm)$) {\textbf{fused}\\\textit{n}\,=\,342};
\node[rec, anchor=north] (sep)   at ($(cross.south)+(-9.15mm,-5mm)$)  {\textbf{separate}\\\textit{n}\,=\,351};
\node[rec, anchor=north] (sonly) at ($(cross.south)+(9.15mm,-5mm)$)   {\mbox{\textbf{spine only}}\\\textit{n}\,=\,89};
\node[rec, anchor=north] (ponly) at ($(cross.south)+(27.45mm,-5mm)$)  {\mbox{\textbf{pelvis only}}\\\textit{n}\,=\,20};
\draw[ln] (cross.south) -- ($(cross.south)+(0,-2.5mm)$);
\draw[ln] ($(fused.north)+(0,2.5mm)$) -- ($(ponly.north)+(0,2.5mm)$);
\foreach \n in {fused,sep,sonly,ponly} \draw[ar] ($(\n.north)+(0,2.5mm)$) -- (\n.north);
\node[proc, anchor=north] (fix) at ($(cross.south |- fused.south)+(0,-5mm)$) {\textbf{correct the source ground truth}\\
  112 spine fixes in 103 cases; 13 pelvic fixes in 12 cases};
\foreach \n in {fused,sep,sonly,ponly} \draw[ln] (\n.south) -- ($(\n.south)+(0,-2.5mm)$);
\draw[ln] ($(fused.south)+(0,-2.5mm)$) -- ($(ponly.south)+(0,-2.5mm)$);
\draw[ar] ($(fix.north)+(0,2.5mm)$) -- (fix.north);
\node[proc, below=of fix] (dense) {\textbf{pseudolabel the missing half, asymmetrically}\\
  a pelvis has no enumeration to get wrong; the 20 pelvis-only spines were corrected by hand};
\draw[ar] (fix.south) -- (dense.north);
\node[ver, anchor=north west] (v3) at ($(dense.south west)+(1mm,-5mm)$) {\textbf{1}};
\node[det, anchor=north west] (d3) at ($(v3.north east)+(2.5mm,-0.5mm)$) {\textbf{femora}, lower thoracic levels, and the \textbf{sacrum}};
\node[ver, anchor=north west] (v4) at ($(v3.south west |- d3.south)+(0,-3mm)$) {\textbf{2}};
\node[det, anchor=north west] (d4) at ($(v4.north east)+(2.5mm,-0.5mm)$) {\textbf{ribs}: M\"oller's rib net unioned with TotalSegmentator, numbered from it; 152 cases reviewed};
\node[ver, anchor=north west] (v5) at ($(v4.south west |- d4.south)+(0,-3mm)$) {\textbf{3}};
\node[det, anchor=north west] (d5) at ($(v5.north east)+(2.5mm,-0.5mm)$) {\textbf{lumbar ribs} as their own class rather than rib~12 (16 cases)};
\node[ver, anchor=north west] (v6) at ($(v5.south west |- d5.south)+(0,-3mm)$) {\textbf{4}};
\node[det, anchor=north west] (d6) at ($(v6.north east)+(2.5mm,-0.5mm)$) {\textbf{surgical instrumentation} as its own classes (11 cases)};
\draw[ar] (dense.south) -- ($(dense.south)+(0,-2.5mm)$) -| (v3.north);
\draw[ar] (v3.south) -- (v4.north); \draw[ar] (v4.south) -- (v5.north); \draw[ar] (v5.south) -- (v6.north);
\node[gate, anchor=north] (qc) at ($(dense.south |- d6.south)+(0,-5mm)$) {\textbf{release gates, in order}\\
  affine and sidedness \textrightarrow\ each rib articulates with the vertebra its number implies
  \textrightarrow\ measures in physiological range};
\draw[ar] (v6.south) -- ($(v6.south)+(0,-2.5mm)$) -| (qc.north);
\node[fin, below=of qc] (rel) {\textbf{802 released records}\\33 with a consensus Castellvi grade};
\draw[ar] (qc.south) -- (rel.north);
\path ([yshift=-2.7pt]current bounding box.south);
\end{tikzpicture}
\caption{How the dataset was built. $n$ counts scans, or corrections, at each step.}
\label{fig:pipeline}
\end{figure}

\subsection{Sources, and why a crosswalk was necessary}\label{sec:sources}
CTSpine1K's 1{,}005 volumes and CTPelvic1K's 1{,}184 come from four and seven collections of
CT obtained for different indications. Only the Cancer Imaging Archive (TCIA) CT colonography
collection (COLONOG)~\cite{colonog,tcia} is common to both, and it is the largest in each: a
subset of a prospective screening trial of asymptomatic patients aged fifty and over, acquired
under one protocol. Every COLONOG patient was scanned prone and supine at a minimum, with 12
exceptions among the 825 patients the sources drew on. CTSpine1K states that one position was
chosen at random by a released script; that script, since deleted but preserved in the
repository's history, writes every series holding twelve or more files to the same filename,
so the volume retained is whichever series the directory walk reached last. No code was
released with CTPelvic1K. Consistent with two independent choices, the sources annotated the
same series for 342 of the 693 patients both annotated (49.4\%), and the released records mix
positions and reconstructions accordingly: 422 supine, 377 prone, three decubitus; 562 at
1.0\,mm, 232 at 1.25\,mm and eight at other slice thicknesses.

Recovering the archived series each annotation was drawn on is nontrivial because the DICOM
attributes that identify an acquisition, the patient, study and series instance identifiers
and the patient position,\cite{dicomps3} have no counterpart in the NIfTI-1
header\cite{nifti1} and are discarded on conversion.\cite{li2016} CTSpine1K's
\texttt{Path.csv} names the TCIA directory each label was converted from, but it carries no
series identifier, names no position for 217 of the 782 annotated records, and for seven
indicates an acquisition other than the one whose bone the label overlaps best.
CTPelvic1K released no imaging and encoded the mapping in its mask filenames,
\texttt{dataset2\_\allowbreak PatientID\_\allowbreak SeriesNumber\_\allowbreak
undocumented\_\allowbreak [tag\_]\allowbreak mask\_4label.nii.gz}. One of the 714 masks names a
patient with no archived series; of the other 713, the first two fields resolve 702 to the
series selected here by bone overlap, five to a \texttt{SeriesNumber} two series share, and six
to one no series carries. The third field matches nothing, and the fourth, on 32 masks, mixes
phenotype (\texttt{sacralization} 7, \texttt{semisacralization} 2,
\texttt{hard\_sacralization} 1), difficulty (\texttt{hard} 8, \texttt{veryhard} 2), image
condition (\texttt{lowdose} 6; \texttt{metal}, \texttt{crop}, \texttt{IntestinalCalculus} once
each) and two unexplained tags (\texttt{dqjoint} 2, \texttt{ydjoint} 1) in one string.

The crosswalk (Fig.~\ref{fig:pipeline}) therefore resolved every annotation, not only the
27.7\% with no stated position, by thresholding each of the patient's volumes at bone
attenuation and keeping the series the mask overlaps most; where a source did name a series,
558 of the 565 CTSpine1K records naming a position and 702 of the 713 CTPelvic1K masks naming
an archived patient agree with it. The result is 342 records whose spine and pelvic labels
land on the same series (\emph{fused}), 351 whose labels land on different series of one
patient (\emph{separate}), 89 with a spine label only and 20 with a pelvic label only. Turning
a patient between prone and supine alters lumbar alignment and the seating of the pelvis
against it~\cite{hasegawa2018}, so a mask drawn on one series does not transfer rigidly to the
other, and while a patient's two annotations lie on different scans no spinopelvic parameter
can be calculated across them and the lumbosacral interface is not one geometry to learn from.

\subsection{Densification, and why it is asymmetric}

A rigid cross-registration of each pelvic mask onto the patient's other acquisition was
attempted, and is released in the code base, but was found wanting. The current
nnU-Net~\cite{nnunet} framework has an \texttt{ignore} label that permitted training on
pelvis-only, spine-only and fused masks together without degrading segmentation at the
lumbosacral junction. A preliminary network was trained in five folds to pseudolabel the
missing pelvises of every volume in the cohort, excluding those with fused masks, using
out-of-fold inference so that no pelvis was pseudolabelled by a model that had seen it during
training. This detail is important because it permitted comparing the accuracy of the model
against the ground-truth radiologist-annotated pelvises: on the manual pelvic labels of each
held-out fold the model reaches a Dice of 0.980 for the sacrum and 0.97 for each hip
(Table~S1). The pelvis is thus radiologist-derived on the 362 fused and
pelvis-only records and pseudolabeled on the other 440 (Fig.~\ref{fig:pipeline}). Those 440 pseudolabelled pelves are close to radiologist quality: a sacrum Dice of $0.980 \pm 0.003$ with no fold below 0.977, and hip Dice of $0.970 \pm 0.016$ and $0.971 \pm 0.015$ with no fold below 0.953, mean the model and the radiologist agree on 97--98\% of the voxels of every pelvic bone on records the model never saw; the per-fold values are released with the data (\texttt{results/pseudolabel\_dice/}). A pelvis carries no enumeration to get wrong,
whereas a pseudolabelled spine must commit to a count, so the 20 pelvis-only spines were
corrected by hand.

\subsection{Ribs}

No public rib dataset annotates the spine with a labeled lumbosacral or thoracolumbar
transitional vertebra, and the two segmenters available for pseudolabeling both
fail~\cite{moller2026,totalsegmentator}.

Möller's binary rib network~\cite{moller2026} works very well when the whole rib is in the FOV,
but fails on a rib even partly outside it, and since ribs grow more caudally angled toward the
thoracolumbar junction, that failure puts a non-negligible amount of bone into the background.
TotalSegmentator~\cite{totalsegmentator} segments ribs both partly and wholly inside the FOV
and assigns each a numeric class, which a binary network cannot, but its ribs stop short of the
costovertebral interface. That blocks the quality-control test confirming that a rib's class
label matches the class of the vertebra it is incident on.

The gaps being complementary, the two were combined (Fig.~\ref{fig:pipeline}): TotalSegmentator
supplied the numbering and the partial-FOV ribs, Möller the ribs wholly inside it and reaching
the vertebra. Their union is a pseudolabel of every rib in the dataset. Rather than review all
of them, a quality-control pipeline triaged the likely errors (Table~\ref{tab:ribqc}) on two
label-based checks: a rib bone carrying two or more labels, and a rib that did not reach the
spine.

The triage identified 152 records. Every one was corrected by a reviewer and 149 finalised by a
second read; the three not finalised are named in the release notes. Neither gate fires on any release record. What is left is
seven ribs split inside the reconstructed field (five carrying a piece of a neighbour, two with
the head apart), two offset ribs and eleven whose medial end stops short of a vertebra.
Reviewers were medical students working through \consortium~\cite{osc2026}, trained against a
written labeling protocol, with every correction attributed to its annotator and a save refused
until the review checks passed. Since a rib is named for the vertebra it articulates with, the
whole-side shifts left after review were renumbered against the vertebrae by that rule, so the
release column is a consistency check on the rule and the pseudolabel column the independent
measure of the pseudolabel and the reviewers.

\begin{table}[!ht]
\caption{Rib checks on the raw pseudolabel and on the release, the same code on both. The
first two gated the review; the rest were measured on every record. Pairs are records /
ribs; a piece is a second component of at least 50 voxels and 15\% of the largest, and a rib
that leaves the reconstructed field and re-enters it is not counted as split.}
\label{tab:ribqc}
\begin{ruledtabular}
\begin{tabular}{@{}l@{\hspace{1.4em}}r@{\hspace{1.4em}}r@{}}
Check & Pseudolabel & Release \\ \colrule
Two numbers on one bone       & 151 / 584 & 0 / 0 \\
Rib detached from the spine   & 37 / 40   & 0 / 0 \\
Rib in two pieces inside the field & 119 / 280 & 6 / 7 \\
Rib offset from its vertebra  & 24 / 170  & 0 / 2 \\
No vertebra within reach      & --- / 38  & --- / 11 \\
Rib on a lumbar body          & 14 / 21   & 0 / 0 \\
\end{tabular}
\end{ruledtabular}
\end{table}

\subsection{Label scheme, and the phenotypes it expresses}\label{sec:anchors}

The scheme is VerSe-native: vertebrae keep their VerSe identifiers (C1--C7 = 1--7,
T1--T12 = 8--19, L1--L6 = 20--25, sacrum 26, coccyx 27, T13 = 28), and every non-VerSe structure
takes a fixed identifier above that range --- hips 30--31, femora 32--33, thirteen ribs per side
34--46 left and 47--59 right, lumbar ribs 60--61 and surgical hardware 62--68.
Identifier 29 is retired and unused, an earlier scheme having given it to a separate S1 class,
and is not reused because renumbering 30--68 to close the gap would rename the hips, femora and
every rib under identifiers consumers are already keyed to. Rib 13 is the true rib of a T13 and
is empty in this release, so a thirteenth thoracic vertebra and a lumbar rib are labeled as the
distinct phenotypes they are.

The sacrum is not sub-divided: it is the source annotation's own outer boundary, no S1 class is
asserted, and the superior end-plate that sacral slope and pelvic incidence are measured from
is fitted to the whole sacrum, isolated by the lowest lumbar body's footprint.

VerSe~\cite{verse2021} carries L6, whose identifier this release keeps, so its L6 \emph{is}
VerSe's. A rib on a lumbar body, however, takes its own class, the one class here with no
published counterpart. A scheme that numbers every rib 1--12 leaves the annotator two bad
options: call it rib~12, which asserts that the vertebra beneath it is thoracic, or discard it.
TotalSegmentator carries ribs 1--12 per side and no such class, and VerSe's T13 is a vertebra
rather than a rib. This release records both: VerSe's T13 (28) with its rib pair (46, 59) for a
thoracic-type thirteenth vertebra, and 60/61 for a lumbar-type body bearing a rudimentary rib. No record in this dataset carries a T13. In each of the sixteen lumbar-rib cases the vertebra
under the rib was named a lumbar body in the source annotation, and the rib class follows by
rule, not judgement: the rib whose head sits on a lumbar body takes 60 or 61. Where a border
vertebra could not be settled, the source reading stands rather than being overridden here.

\begin{figure*}[t]
\centering
\includegraphics[width=\linewidth,keepaspectratio]{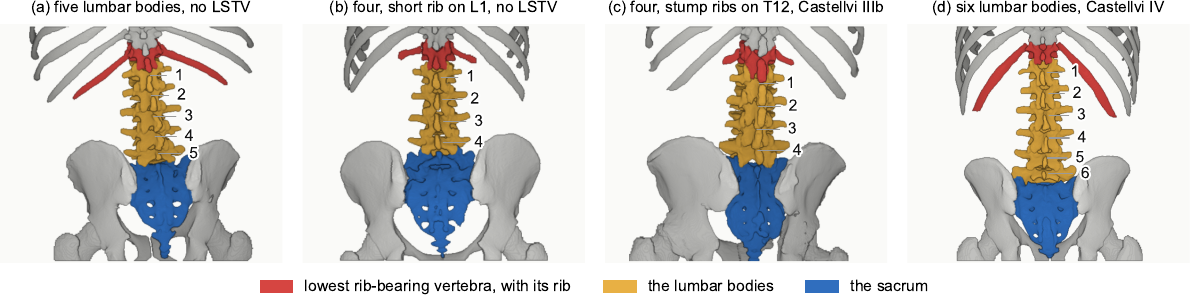}
\caption{Four phenotypes at the lumbar borders, aligned on the sacrum. Red, lowest rib-bearing
vertebra and its rib; blue, sacrum; yellow, the lumbar bodies. Cases 0704, 0428, 0094,
0005, each the most ordinary member of its phenotype by rule. (b) and (c) present the same
upper border, a short rib pair under a full pair, and were named differently.}
\label{fig:anchors}
\end{figure*}

Figure~\ref{fig:anchors} shows the four phenotypes the scheme has to express, and why a count
does not settle any of them. (a) is the ordinary column, five lumbar bodies and no transitional
vertebra. (b) and (c) present the same upper border --- a rib pair of about 40\,mm beneath a
full pair, above four lumbar bodies --- and were named differently: a lumbar rib on L1 in (b),
stump twelfth ribs on T12 in (c). Nothing in the rib decided that, and the two populations
barely separate on rib length alone, the sixteen ribs labeled lumbar running a median 45\,mm
and 0.32 of the rib above them against 38\,mm and 0.28 for the 98 labeled stump twelfth ribs.
What decided it was the convention of five lumbar vertebrae, which leaves exactly one place for
the anomaly once the lower border is read, transitional in (c) and not in (b). (d) is the
sixth lumbar body. A scheme carrying only ribs 1--12, five lumbar identifiers and no sacrum
cannot write down (b), (c) and (d) as the distinct things they are; this one can, which is what
makes them separable data rather than three readings of one count.

So the name a border vertebra carries in this release is a reader's, not a measurement's. Every
vertebra keeps the identifier its radiologist-sourced annotation gave it, corrected only where
that source was internally wrong, and no morphometric criterion was applied to overrule a
reading: a T12 and an L1 at this border are separated here by the eye that read them. That is
the convention every public collection follows, and it is worth stating rather than assuming,
because the measurements released beside these labels are what a morphometric criterion would
be built from, and some of these labels may not survive one. A user testing such a criterion
should treat the border labels as the reference standard they are --- expert reads --- and not
as ground truth independent of the reader.

Of the sixteen, thirteen are bilateral and three unilateral, all on the right, too few to speak
to side preference.

Eighteen records carry an L6. Seventeen of them carry a consensus Castellvi grade;
the eighteenth was not graded, being the one instrumented record, pulled out of the normal
review flow while its merged lumbar labels were rebuilt by hand and never returned to the
readers. The source transitional label (pelvic where CTPelvic1K flagged one, otherwise the lumbar count) reads lumbarization in fourteen, sacralization in two and semi-sacralization in one, and is unremarkable in the ungraded record. \begin{sloppypar} Both counts, and the manifest fields that report them (\texttt{has\_l6},
\texttt{has\_lumbar\_rib}, \texttt{n\_lumbar\_labels}, \texttt{lumbar\_rib\_side}), are
computed from the released label volumes by counting identifiers 25 and 60--61.
\end{sloppypar}

\subsection{Computational tools}\label{sec:tools}

Every step in this pipeline is performed by open-source code archived with the
dataset, together with the reference loader, the label scheme and the quality-control scripts
that generated the figures and tables in this manuscript.

Segmentation of the femora, the sacral sub-division and the
per-level rib numbering used TotalSegmentator~\cite{totalsegmentator} (\texttt{total} task,
release 2.x) on a single graphics processing unit. The binary rib network is
M\"oller's~\cite{moller2026} released weights, applied through the nnU-Net v2
framework~\cite{nnunet}. The pelvic completion for records whose pelvis was absent used a five-fold nnU-Net v2 ensemble
applied out-of-fold, so no record is completed by a model that has seen it. The five fold
checkpoints, with their nnU-Net plans and dataset descriptor, are released alongside the
data (\ifreview
\else\url{https://huggingface.co/OpenSpineConsortium/spinopelvic-seg-checkpoints}, trained with \url{https://github.com/Gregory-Schwing-MD-PhD/spinesurg-ct-nnunet} at commit 50b209f\fi). Image handling throughout used NiBabel and SciPy; no image is resampled or reoriented when a
label is written, so every label shares its image's grid and affine exactly.

Dimensions and spinopelvic parameters come from code released with the
dataset. A sacral plate whose fitted normal lies more than 60\si{\degree} off the cranial axis is
rejected and the case reported missing, since such a fit has found the near-vertical anterior
face of the promontory, not the endplate. Gating on the geometry rather than on the
value leaves unusual patients in. \Toolkit{} is released separately
(\ifreview\else\url{https://github.com/OpenSpineConsortium/OpenSpineToolkit}\fi)
as a reusable implementation measuring in each vertebra's own frame. No open-source
implementation computing these parameters from CT segmentations was found elsewhere.

A large language model (Claude, Anthropic) assisted
with writing the analysis and figure code and with manuscript outlining and refinement. It was
not used to generate, annotate or interpret imaging data; every number reported here is computed by the released code from the
released volumes. All content was independently reviewed, verified, and interpreted by the authors.

\subsection{Validation}
\label{sec:validation}

Validation has three levels. Geometry is checked first, because a measurement taken on a
structure with a geometric fault still returns a plausible-looking number.

Every case is checked for label geometry agreeing with its CT in
shape, affine and voxel spacing; for identifiers within the published scheme; for a non-empty
label; and for sided structures falling on the correct sides, with the left--right axis read
from the affine rather than assumed. 

The check compares \emph{each sided pair separately}. Testing the ribs alone passed all 802
records, four of which carried a \texttt{left\_hip} label on the patient's right side, and
pooling every sided structure into one test still missed one of those four, because a large
correctly-sided structure masks a smaller transposed one. A gate that passes everything is not
evidence of a clean corpus until it has been shown capable of failing.

Each rib is matched to its nearest vertebra and compared
against its corresponding vertebra. Of 11{,}560 ribs, 5{,}796 are not evaluable
because the expected vertebra lies outside the FOV and 11 have no vertebra within
the anchor distance. Of the 5{,}753 evaluable, 5{,}751 match and two (0.035\%) are offset by
one level. The denominator is stated because quoting the two offsets against all 11{,}560
ribs would halve the rate by counting ribs the check never examined. No case is misnumbered,
which is a different call: a rib is offset when the vertebra it reaches is not the one its
number implies, whereas a case is misnumbered only when three or more of its ribs are offset by
the same amount, the signature of a side numbered from the wrong end. One or two isolated
offsets are likelier a segmentation artefact, and that is what these are, single ribs in two
records, both truncated by the edge of the field.

Of the 5{,}753 evaluable ribs, \emph{zero} are assigned to a lumbar vertebra, despite 16 records
carrying one, because a lumbar rib takes class 60 or 61 and never enters the numbered set the
test examines. A numbered rib landing on a lumbar vertebra would be the labeling error this
dataset exists to expose.

Pelvic incidence is the strongest of these checks
because it is a morphological property rather than a posture. This cohort's mean agrees to
within a tenth of a degree with the automated supine CT series of Veilleux \emph{et
al.}~\cite{veilleux2020} and sits below the standing figure, the direction reported for subjects
imaged both ways.\cite{leeliu2022} Sacral slope and pelvic tilt are postural, and read within a tenth of a degree of Veilleux's asymptomatic subjects. A second supine CT series reports
53.4\si{\degree}, 34.1\si{\degree} and 19.2\si{\degree}, within four degrees of this cohort on all
three.\cite{hasegawa2018} That residual is measured against 24 subjects, and both comparator
series are supine, so posture cannot account for it; it tracks cohort and not construction.

\subsection{Castellvi typing}

Every record whose lumbar vertebra count differs from five carries a Castellvi
grade~\cite{castellvi1984} in the release, 33 cases typed I--IV with the $a$/$b$
unilateral--bilateral qualifier. The grade and the count are \emph{different axes}, the count
being how many lumbar bodies the column holds. Grade IIIb occurs here at lumbar counts of four,
five and six alike, and seven of the 33 graded cases carry a normal count of five.

Two radiology resident physicians graded every case independently and resolved their six
disagreements by consensus; both reads and the consensus are in the manifest. The distinction
they disagreed on, type II (articulation) against type III (bony fusion)~\cite{koninwalz2010}, is
the one that matters clinically, and four of the six disagreements were III against II.

\begin{table*}[t]
\caption{Spinopelvic measures, mean $\pm$ SD over the 756 of 802 records passing the geometric
gates, against two supine CT series that report means.\cite{veilleux2020,hasegawa2018} Standing values are
55.0, 41.0, 13.0 and 60.0\si{\degree}.\cite{vialle2005} The end-plate is fitted to the whole
sacrum, and the plate gate rejects 44 records. Pelvic incidence and
tilt are right-skewed (0.25 and 0.33), so their medians---51.5 and 15.0\si{\degree}---sit below
these means and the modal peak sits below both.}
\label{tab:spinopelvic}
\begin{ruledtabular}
\maxwidth{%
\begin{tabular}{lccc}
Measure & This dataset (supine, $n=756$) & Veilleux ($n=200$) & Hasegawa ($n=24$) \\ \colrule
Pelvic incidence & 52.1 $\pm$ 11.1\si{\degree} & 52.1 & 53.4 \\
Sacral slope & 36.6 $\pm$ 10.6\si{\degree} & 36.5 & 34.1 \\
Pelvic tilt & 15.5 $\pm$ 7.9\si{\degree} & 15.6 & 19.2 \\
Lumbar lordosis & 51.7 $\pm$ 13.3\si{\degree} & --- & --- \\
PI $-$ LL mismatch & 0.3 $\pm$ 10.3\si{\degree} & --- & --- \\
\end{tabular}
}
\end{ruledtabular}
\end{table*}

\subsection{Surgical hardware, and why a threshold is not enough}
\label{sec:hardware}

Metal is trivial to detect in CT and hard to interpret, and the difference matters here because
\textbf{to a distance measurement an iatrogenic fusion is indistinguishable from a congenital
one}: a cage-bridged interspace reads as ``no gap'' exactly as a fused transitional vertebra
does.

A 2500\,HU threshold inside a shell around the labeled skeleton proposes candidates, and every
candidate is confirmed or rejected by manual review, because attenuation alone cannot separate
an implant from contrast, calcification or reconstruction artifact --- all of them saturate.
Site and volume can: every confirmed implant measures 2,586\,mm$^3$ or more and every rejection
1,768\,mm$^3$ or less, and the confirmed object is typed from its site and shape. Identifiers
62--65 name spinal instrumentation (generic, cage, screw-and-rod, plate), to which this release
adds \textbf{66 arthroplasty}, \textbf{67 sacroiliac screw} and \textbf{68 osteosynthesis}.
Eleven records carry hardware: eight arthroplasties, one osteosynthesis, one sacroiliac
fixation and one pair of interbody cages, each shown in Fig.~S1.

An implant lies inside the bone label that surrounds it, so the hardware class is taken
\emph{out} of that label rather than added beside it, reclaiming 1,538,852\ voxels across
the eleven records: a user measuring bone gets bone, and one measuring the implant gets the
implant. In eight of the eleven, the femoral head that pelvic incidence and tilt are measured
from is an implant.

\section{Data Format and Usage Notes}\label{sec:format}

Each record is a gzipped NIfTI image/label pair sharing an identical $4\times4$ affine,
canonicalized to PIR, requiring no resampling. Masks are NIfTI rather than DICOM, which is what
volumetric segmentation tooling expects.

Splits are patient-grouped and LSTV-stratified five-fold, frozen and shipped with the data
as \texttt{splits\_5fold.json}. Every record is a validation record in exactly one of the
five folds, so the five validation sets together cover the whole cohort. The release
therefore supports cross-validation, but it contains no set of records held back from all
five folds. A reader who wants to quote one accuracy figure on data a model has never seen
must set records aside before training, keep them out of every fold, and say which records
those were.

Stratification is on the transitional subtype rather than a binary LSTV flag. Across all
802 records the source transitional label reads lumbarization in 14, sacralization in 17
and semi-sacralization in two. Eighteen records carry a sixth lumbar-type vertebra, nine
carry only four lumbar identifiers, and 33 have a consensus Castellvi grade.

The strata themselves are read from the source transitional label, not from the lumbar
count: the released file holds 768 normal, 15 sacralization, 15 lumbarization, two
semi-sacralization and two whose two sources disagree. The two strata of 15 put three
records in every validation fold; the two strata of two cannot appear in every fold at all.
That is the most stratification can guarantee with these numbers, and it is why a
fold-level metric on the rare classes carries an error bar far wider than its own decimal
places. A reader who wants folds balanced on the label volumes instead --- on which records
carry a sixth lumbar identifier, say --- must regenerate them, because the shipped splits do
not encode that.

Per-field record counts across the manifest are given in Table~S2.

The archive of record is Zenodo (\doiurl{\datasetdoi}). Code and documentation are on GitHub at
the tagged release commit; any model-hub copy is a convenience mirror, not the archive.

\section{Descriptive Analysis}
\label{sec:descriptive}


The cohort is a colorectal cancer screening population of 802 records: 393 female, 345 male,
11 carrying DICOM's \emph{other} value and 53 with no sex field, with age present for 709
(median 59, range 50--89). All 802 are counted in totals, and the 64 who are not recorded as female or male --- the 11
carrying DICOM's \emph{other} value and the 53 with no sex field --- are excluded from
sex-stratified measures, which therefore rest on 738 records.

Five lumbar bodies is typical; four and six are where transitional anatomy sits, and which is
present does not by itself determine what to call it. The lowest-rib length ratio is bimodal,
its lower mode lying below 0.33 of the rib above. Its lower mode marks a stump twelfth rib in 98 of the 789 records with a measurable pair
(12.4\%), and a lumbar rib is present in 16 (2.0\%), against 12.6\% and 1.5\% in a
whole-spine CT series that reports both~\cite{nagata2025}. That series also reports that the two rib anomalies travel with the
lumbosacral border --- stump twelfth ribs with sacralization, lumbar ribs with lumbarization
--- and it is that pairing of a rib anomaly with a transitional junction, rather than the two
rib types occurring together, that is testable here. It is testable only where the junction
was read, and only within those 789: they hold 14 of the 17 source-labeled sacralizations, and stump ribs accompany four of those 14 and 94 of the other 775 (odds ratio 2.9, $p=0.08$, Fisher's exact test). No lumbarization case carries a lumbar rib.

Pelvic incidence sits on the standing reference, as a morphological property should, while
sacral slope and pelvic tilt fall above and below the published CT values (Fig.~S2); Table~\ref{tab:spinopelvic} places all three against two
series.


Every scan already contains the information needed to measure vertebral bone density, so it
can be read off a study acquired for another indication without exposing the patient to any
further radiation.\cite{pickhardt2013} L1 trabecular attenuation is reported here at the
published standard site.

\section{Potential Applications and Limitations}\label{sec:applications}

The principal use is the \emph{spinopelvic interface}. Spine and pelvis carry
radiologist-sourced annotation on one coordinate frame in 802 records, so sacral morphology is
measurable against the lumbar column rather than in isolation: the sacral endplate and its
slope, the alar corridors constraining S2-alar-iliac and iliac screw trajectories, and the
L5--S1 geometry a lumbosacral fusion must cross. Pelvic incidence is derived from the segmented
sacrum and femoral heads, so it is reproducible from the labels themselves. No trajectory can
be planned across a junction whose segments cannot be named, so the transitional layer serves
that use rather than competing with it.

\emph{A proxy for the preoperative view.} We are not aware of a public preoperative lumbar CT cohort, which makes this release the
closest available stand-in for the view that lumbar cases are planned on. This dataset contains the T12-to-S1 span, including the lowest ribs and the pelvis. 377 of its 802 records were acquired prone, the position of posterior lumbar surgery, and it contains classes for the anomalies that make level identification fail. The cohort is a screening population at colonography dose, not a surgical series, so it supports method development for anatomical level identification and instrumentation geometry rather than validation of a preoperative planning decision.

\subsection{Three uses the released measurements support}

Wrong-level spine surgery runs at roughly one in 3{,}110 procedures, and among the predominant
contributing factors is the transitional anatomy this cohort was assembled
around.\cite{mody2008,epstein2021} Per-level body height, canal width, end-plate width,
transverse-process span and wedge ratio are released across all 802 records, so a model can be
trained on the shape of a vertebra rather than on its position in a count.

\begin{figure*}[t]
\centering
\includegraphics[width=\linewidth,keepaspectratio]{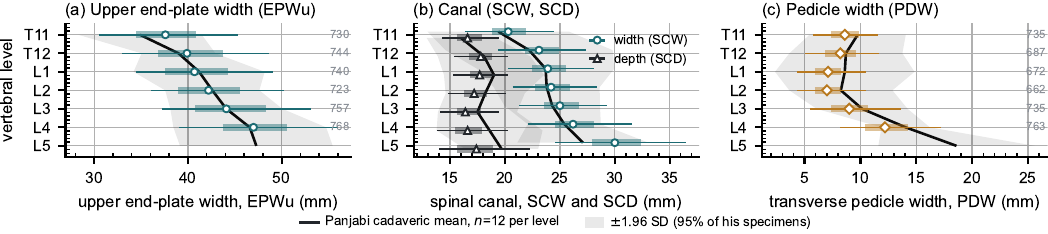}
\caption{Morphometry by level, T11--L5: median, interquartile range, 5th--95th
percentile, $n$ at right. (a) End-plate width EPWu. (b) Canal width SCW and depth SCD.
(c) Pedicle width PDW, both sides averaged. L5 is withheld in (a) and (c); see text.
Reference is Panjabi's cadaveric series\cite{panjabi1991,panjabi1992}, keyed in (a), its
SD recovered from the reported SEM; T11--T12 EPWu is digitised from his Figure 4A.}
\label{fig:levelatlas}
\end{figure*}

The values a surgeon uses for endplate
and canal dimensions and pedicle width come from cadaveric series of ten to thirty
specimens\cite{panjabi1991,panjabi1992,berry1987} redrawn in the textbooks as one curve per
level\cite{benzel2015}, and a standard error describes the mean, not the spread of
individuals. The same measures are given here from 802 records with the distribution attached
(Fig.~\ref{fig:levelatlas}), reproducing the classical pedicle widening into the lower lumbar
spine. Both layers of that figure are population intervals, not confidence intervals, and that
is what makes their widths comparable: a confidence interval bounds a parameter, a population
interval bounds individuals,\cite{hahn2017} and at $n=12$ the two differ by $\sqrt{12}$.

Patient-specific finite-element modeling for surgical alignment
planning and implant design\cite{fea2026} needs spine, pelvis and femoral heads in one
frame, which is provided by this release. 351 patients in this dataset give a within-patient
postural change to check a prediction against.

\label{sec:limitations}Thoracic ground truth for this release is
FOV limited and does not reach T1 (Fig.~S3). In addition, the postural angles measured are supine. While pelvic incidence is not postural, it is modality-sensitive, and the measurements read below
radiography for the same subjects.\cite{leeliu2022} End-plate and pedicle width are
withheld at L5 because the transverse processes arise in front of the canal wall the body is cut at. Those measurements along with the spinopelvic angles are
the provisional part of this release. The planned revision takes the vertebral body and the sacral plate from a substructure
label, and the version DOI carries it. The cohort is also one protocol of a colorectal
screening population aged 50 and over, so its distributions should not be read as those of a
surgical series.

Label strength varies by structure, and the release does not average over it.
Vertebral labels derive from radiologist-supervised source annotations. Where those source
labels were wrong they were corrected by trained medical-student reviewers working to a
written protocol, with any case whose junction they could not settle escalated to a
radiologist rather than overwritten. Pelvic labels on records that lacked one are
pseudolabeled.

The ribs are pseudolabels, and their review was triaged rather than exhaustive: a rule
selected the records most likely to be wrong and those were corrected, so what the release
can promise is that the specific failures that rule detects are absent, not that every rib
in every record has been inspected. The transitional labels come from two independent
sources and are not uniformly adjudicated, which is why the measures reported here do not
depend on the count; the Castellvi grades are a consensus of two radiology resident
physicians. The sacrum is from the source annotation's own outer
boundary and there is no sub-division of it asserted, so a user that needs a first sacral
segment must delineate one.

Sacrum, hips and femora are present in all 802 records; one lacks T12, outside the FOV. Nine
carry no L5 identifier because their source annotation counted four lumbar vertebrae; all nine
are Castellvi IIIb, the fused segment is delineated with the sacrum, and the manifest names
them. No shape-based classifier is attempted at the lumbosacral junction, where 15 records in
each rare stratum leave three per fold; the 0.990 area under the curve above is the
thoracolumbar boundary, not this one.

\section{Discussion}

VerSe~\cite{verse2021,versedata2021} comes closest in intent to this dataset, but carries no
sacral mask for the structure a Castellvi grade describes, which this release segments as L6
or as sacrum, and has neither the pelvis and femora a spinopelvic measurement needs nor prone
acquisitions. What this release adds
over TotalSegmentator~\cite{totalsegmentator} is the classes for anatomical anomalies, not necessarily more
anatomy per scan.

Applying the released weights to VerSe would add the sacrum and the
anomaly classes to a bone-kernel collection. A Castellvi read of the whole cohort
would test the co-occurrence of rib and lumbosacral anomalies at full power and supply the
labels a lumbosacral classifier needs. A class label from substructure
segmentation\cite{spineps2025} would measure the body without the canal cut that withholds
L5 in this dataset.

\section{Conclusion}

CTSpinoPelvic1K places the radiologist-derived spine and pelvis and the pseudolabeled per-level
ribs and femora on one validated coordinate frame in 802 records, giving the anatomical
anomalies that make lumbar numbering ambiguous classes of their own.
Because every vertebra is measured in its own right rather than by its place in a sequence, a
level can be named from the morphology in a field of view that does not allow the
conventional count downward from C2. This release demonstrates that at the thoracolumbar
border and leaves it open at the lumbosacral one, where it supplies the classes and the
measurements to settle the question but not yet the number of transitional records a
classifier would need. 

\section*{Supporting Information}
Supporting information is available online and is not part of the article PDF.
Fig.~S1, the instrumentation gallery, one exemplar of each implant class drawn through
bone. Fig.~S2, the spinopelvic measures against standing reference bands. Fig.~S3, the
records carrying each vertebral level. Table~S1, the pelvic pseudolabeller's held-out
fold Dice. Table~S2, data types and metadata and the records populating each.
Table~S3, the per-identifier label census.

\section*{Data Availability}
The release described here is \datasetversion{}, permanently archived at
\doiurl{\datasetversiondoi}; the concept DOI \doiurl{\datasetdoi} resolves to the
current version.
The archive carries the loader, the label scheme and the quality-control tables under an
open-source licence. The repository additionally carries the extraction and figure code,
the reference table with a citation on every row, the values plotted in
Fig.~\ref{fig:levelatlas}, and a recipe for running them in order. \ifreview
\else Code:
\url{https://github.com/OpenSpineConsortium/CTSpinoPelvic1K} (article materials under
\texttt{paper/mpda/}); convenience mirror:
\url{https://huggingface.co/datasets/OpenSpineConsortium/CTSpinoPelvic1K}; measurement
toolkit: \url{https://github.com/OpenSpineConsortium/OpenSpineToolkit}.\fi

\section*{Acknowledgements}
Acknowledgements are given on the title page, per the journal's double-anonymized review
policy. This work received no funding.

\section*{Conflict of Interest}
The authors have no relevant conflicts of interest to disclose.

\section*{Ethics}
This work uses publicly available, de-identified imaging and annotations derived from it.
The authors' institutional review board determined that it is not human participant
research under 45 CFR 46 and requires no oversight.

\ifsupplement
\clearpage
\onecolumngrid
\begin{center}{\large\bfseries Supporting Information}\end{center}
\setcounter{figure}{0}\renewcommand{\thefigure}{S\arabic{figure}}
\setcounter{table}{0}\renewcommand{\thetable}{S\arabic{table}}


\begin{figure}[htbp]
\centering
\includegraphics[width=\linewidth,keepaspectratio]{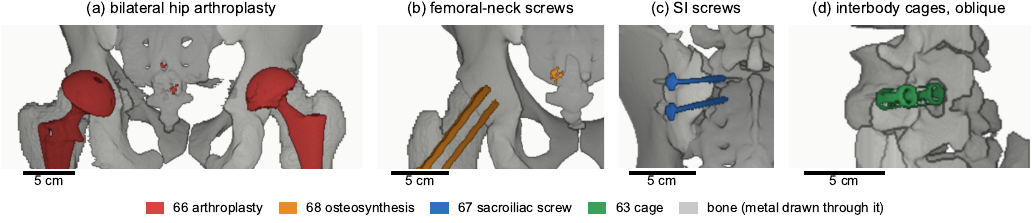}
\caption{Instrumentation in the release, one exemplar each, drawn through bone. (a) Hip arthroplasty, 66, $n=8$. (b) Femoral-neck screws, 68, $n=1$. (c) Sacroiliac screws, 67, $n=1$. (d) Interbody cages, 63, $n=1$. Bar, 5\,cm.}
\label{fig:hardware}
\end{figure}

\begin{figure}[htbp]
\centering
\includegraphics[width=\linewidth,keepaspectratio]{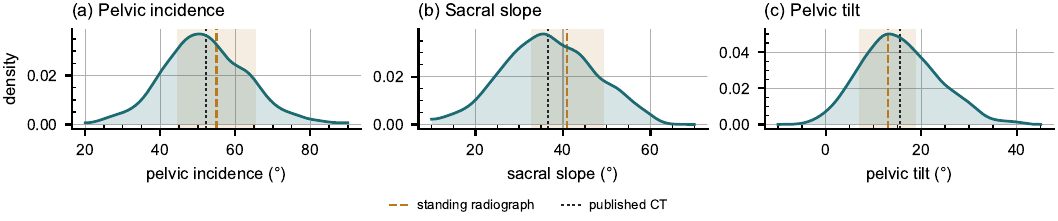}
\caption{Spinopelvic measures against standing reference bands (mean $\pm$
SD)~\cite{vialle2005}.}
\label{fig:validation}
\end{figure}

\begin{figure}[htbp]
\centering
\includegraphics[width=\linewidth,keepaspectratio]{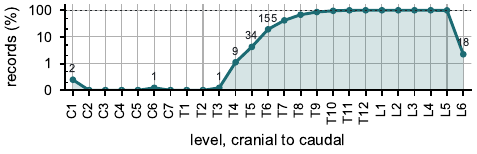}
\caption{Records carrying each vertebral level, cranial to caudal. Counts per identifier
in Table~S1.}
\label{fig:fov}
\end{figure}

\clearpage

\begin{table}[htbp]
\caption{Held-out fold Dice of the pelvic pseudolabeller against manual CTPelvic1K labels,
mean $\pm$ SD over the five folds at each fold's selected checkpoint (nnU-Net validation
Dice, ignore regions excluded; \texttt{results/pseudolabel\_dice/} in the repository).}
\label{tab:pseudodice}
\begin{ruledtabular}
\begin{tabular}{lc}
Structure & Dice \\ \colrule
Sacrum    & 0.980 $\pm$ 0.003 \\
Left hip  & 0.970 $\pm$ 0.016 \\
Right hip & 0.971 $\pm$ 0.015 \\
\end{tabular}
\end{ruledtabular}
\end{table}

\begin{table}[htbp]
\caption{\label{tab:completeness}Data types and metadata, and the records populating
each, from the released manifest.}
\begin{ruledtabular}
\begin{tabular}{lrr}
Field & Records & Available \\
\hline
Identifiers and paths, configuration, match type, image/label alignment check & 802 & 100.0\% \\
Patient position, tube potential, section thickness, kernel & 802 & 100.0\% \\
Transitional label and class, annotation origin, instrumentation flags & 802 & 100.0\% \\
Spine series identifier; bone-fraction check & 782 & 97.5\% \\
Scanner manufacturer and model & 768 & 95.8\% \\
Sex & 749 & 93.4\% \\
Age & 709 & 88.4\% \\
Pelvic series identifier & 713 & 88.9\% \\
Castellvi grade (consensus of two readers) & 33 & 4.1\% \\
\end{tabular}
\end{ruledtabular}
\end{table}

\begin{table}[htbp]
\caption{Every identifier populated in the release and the number of the 802 records
carrying it, counted from the released label volumes. The identifier space runs 0--68
with one hole at the retired 29; 46 and 59 (the thirteenth rib pair) occur in no record.}
\begin{ruledtabular}
\begin{tabular}{lrr}
id\ class & records & \% \\ \colrule
1 C1 & 2 & 0.25 \\
6 C6 & 1 & 0.12 \\
10 T3 & 1 & 0.12 \\
11 T4 & 9 & 1.12 \\
12 T5 & 34 & 4.24 \\
13 T6 & 155 & 19.33 \\
14 T7 & 335 & 41.77 \\
15 T8 & 545 & 67.96 \\
16 T9 & 688 & 85.79 \\
17 T10 & 765 & 95.39 \\
18 T11 & 798 & 99.50 \\
19 T12 & 801 & 99.88 \\
20 L1 & 802 & 100.00 \\
21 L2 & 802 & 100.00 \\
22 L3 & 802 & 100.00 \\
23 L4 & 802 & 100.00 \\
24 L5 & 793 & 98.88 \\
25 L6 & 18 & 2.24 \\
26 sacrum & 802 & 100.00 \\
30 left\_hip & 802 & 100.00 \\
31 right\_hip & 802 & 100.00 \\
32 femur\_left & 802 & 100.00 \\
33 femur\_right & 802 & 100.00 \\
36 rib\_left\_3 & 5 & 0.62 \\
37 rib\_left\_4 & 65 & 8.10 \\
38 rib\_left\_5 & 316 & 39.40 \\
39 rib\_left\_6 & 635 & 79.18 \\
40 rib\_left\_7 & 780 & 97.26 \\
41 rib\_left\_8 & 800 & 99.75 \\
42 rib\_left\_9 & 802 & 100.00 \\
43 rib\_left\_10 & 802 & 100.00 \\
44 rib\_left\_11 & 802 & 100.00 \\
45 rib\_left\_12 & 787 & 98.13 \\
49 rib\_right\_3 & 5 & 0.62 \\
50 rib\_right\_4 & 62 & 7.73 \\
51 rib\_right\_5 & 310 & 38.65 \\
52 rib\_right\_6 & 624 & 77.81 \\
53 rib\_right\_7 & 773 & 96.38 \\
54 rib\_right\_8 & 800 & 99.75 \\
55 rib\_right\_9 & 801 & 99.88 \\
56 rib\_right\_10 & 801 & 99.88 \\
57 rib\_right\_11 & 802 & 100.00 \\
58 rib\_right\_12 & 788 & 98.25 \\
60 rib\_left\_lumbar & 13 & 1.62 \\
61 rib\_right\_lumbar & 16 & 2.00 \\
63 hardware\_cage & 1 & 0.12 \\
66 hardware\_arthroplasty & 8 & 1.00 \\
67 hardware\_si\_screw & 1 & 0.12 \\
68 hardware\_osteosynthesis & 1 & 0.12 \\
\end{tabular}
\end{ruledtabular}
\end{table}

\fi

\end{document}